\documentclass[10pt,twocolumn,twoside]{IEEEtran}

\usepackage{cite}
\ifCLASSINFOpdf
   \usepackage[pdftex]{graphicx}
   \graphicspath{{../pdf/}{../jpeg/}}
   \DeclareGraphicsExtensions{.pdf,.jpeg,.png}
\else
   \usepackage[dvips]{graphicx}
   \graphicspath{{../eps/}}
   \DeclareGraphicsExtensions{.eps}
\fi

\usepackage{subfigure}

\usepackage[cmex10]{amsmath}

\usepackage{amsfonts, amsmath, amssymb, epsfig, graphicx, psfrag, subfigure, cite}
\usepackage{itrans}

\newtheorem{proposition}{Proposition}
\newtheorem{theorem}{Theorem}

\begin{document}
\title{Multi-object Classification via Crowdsourcing with a Reject Option}

\author{Qunwei~Li,~\IEEEmembership{Student~Member,~IEEE}, Aditya~Vempaty,~\IEEEmembership{Member,~IEEE}, Lav~R.~Varshney,~\IEEEmembership{Senior~Member,~IEEE}, and Pramod~K.~Varshney,~\IEEEmembership{Fellow,~IEEE}%
\thanks{This work was supported in part by ARO under Grant W911NF-14-1-0339 and by NSF under Grant IIS-1550145.}
\thanks{Q.\ Li and P.~K.\ Varshney are with the Department of Electrical Engineering and Computer Science, Syracuse University, Syracuse, NY 13244 USA
(e-mail: qli33@syr.edu; varshney@syr.edu).}
\thanks{A.\ Vempaty was with the Department of Electrical Engineering and Computer Science, Syracuse University, Syracuse, NY 13244 USA. He is now with the IBM Thomas J.\ Watson Research Center, Yorktown Heights, NY 10598 USA (e-mail: avempat@us.ibm.com).}
\thanks{
L.~R.\ Varshney is with the Department of Electrical and Computer Engineering and with the Coordinated Science Laboratory, University of Illinois at Urbana-Champaign, Urbana, IL 61801 USA (e-mail: varshney@illinois.edu).}% <-this % stops a space
}

\maketitle
\begin{abstract}
Consider designing an effective crowdsourcing system for an $M$-ary classification task. Crowd workers
complete simple binary microtasks whose results are aggregated to give the final result. We consider
the novel scenario where workers have a reject option so they may skip microtasks when they are unable or choose not to respond. For example, in mismatched speech transcription, workers who do not know the language may not be able to respond to microtasks focused on phonological dimensions outside their categorical perception. We present an aggregation approach using a weighted majority voting rule, where each worker's response is assigned an optimized weight to maximize the crowd's classification performance. We evaluate system performance in both exact and asymptotic forms. Further, we consider the setting where there may be a set of greedy workers that complete microtasks even when they are unable to perform it reliably. We consider an oblivious and an expurgation strategy to deal with greedy workers, developing an algorithm to adaptively switch between the two based on the estimated fraction of greedy workers in the anonymous crowd. Simulation results show improved performance compared with conventional majority voting.
\end{abstract}
\begin{IEEEkeywords}
Classification, crowdsourcing, distributed inference, information fusion, reject option
\end{IEEEkeywords}

\section{Introduction}

\IEEEPARstart{O}{ptimization} problems that arise in communication, computation, and sensing systems have driven signal processing advances over
the last several decades.  Now, engineered social systems such as crowdsensing, crowdsourcing, social production, social networks, and data analytics deployments
that interact with people are becoming prevalent for all of these informational tasks.  Yet, they are often constructed in an ad hoc manner.
Advances in signal processing theory and methods to optimize these novel human-oriented approaches to signal acquisition and processing are needed \cite{PoorCKSW2014}.

Crowdsourcing has particularly attracted intense interest  \cite{TapscottW2006,Howe2008Crowdsourcing,TapscottW2010,Yuen2011Survey,Bollier2011,Hossfeld2014best,hosseini2014four} as a new paradigm for signal processing tasks such as handwriting recognition, paraphrase acquisition, speech transcription, image quality assessment, and photo tagging \cite{paritosh2011computer,kamar2012combining,burrows2013paraphrase,7052378,SprugnoliMFGBPGB2013,HoßfeldHKHGKT2014,RibeiroFN2011,RibeiroFCS2011}, that essentially all boil down to the classical statistical inference problem of $M$-ary classification.  Unfortunately, the low quality of crowdsourced output remains a key challenge \cite{IpeirotisPW2010,allahbakhsh2013quality,mo2013cross}. 

Low-quality work may arise not only because workers are insufficiently motivated to perform well \cite{Varshney2012e,hirth2013analyzing}, but also because workers may lack the skills to perform the task that is posed to them \cite{VempatyVV2014}. Decomposing larger tasks into smaller subtasks for later aggregation allows workers lacking certain skills to contribute useful information \cite{KargerOS2011b,VempatyVV2014}, by polarizing the lack of skill into a subset of subtasks.

As an illustrative example of lack of skill, consider the problem of mismatched crowdsourcing for speech transcription, which has garnered interest in the signal processing community \cite{Hasegawa-JohnsonCJV2015,jyothi2015acquiring,VarshneyJH2016,LiuJTMSKHK2016,JyothiH2015b,ChenHC2016,KongJH2016}.  The basic idea is to use crowd workers to transcribe a language they do not speak, into nonsense text in their own native language orthography.  There are certain phonological dimensions, such as aspiration or voicing, that are used to differentiate phonemes in one language but not another \cite{VarshneyJH2016}.  Moreover due to categorical perception that is learned in childhood, workers lacking a given phonological dimension in their native language may be unable to make relevant distinctions.  That is, they lack the skill for the task.

\begin{figure}
   \centering
   \includegraphics[width=3.5in]{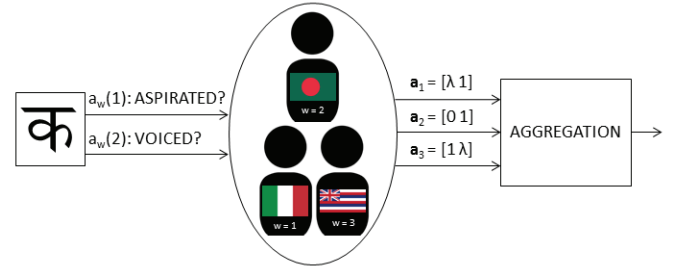} % requires the graphicx package
   \caption{An illustrative example of the proposed crowdsourcing framework.}
   \label{example}
\end{figure}

Fig.~\ref{example} depicts the task of language transcription of Hindi. Suppose the four possibilities for a velar stop consonant to transcribe are $R=\{${\fransdvng k}, {\fransdvng K}, {\fransdvng g}, {\fransdvng G}$\}$. The simple binary question of ``whether it is aspirated or unaspirated'' differentiates between $\{${\fransdvng K}, {\fransdvng G}$\}$ and $\{${\fransdvng k}, {\fransdvng g}$\}$, whereas the binary question of ``whether it is voice or unvoiced'' differentiates between $\{${\fransdvng g}, {\fransdvng G}$\}$ and $\{${\fransdvng k}, {\fransdvng K} $\}$. Now suppose the first worker is a native Italian speaker.  Since Italian does not use aspiration, this worker will be unable to differentiate between $\{${\fransdvng k}$\}$ and $\{${\fransdvng K}$\}$, or between $\{${\fransdvng g}$\}$ and $\{${\fransdvng G}$\}$.  It would be of benefit if this worker would specify the inability to perform the task through a special symbol $\lambda$, rather than guessing randomly. Suppose the second worker is a native Bengali speaker; since this language makes a four-way distinction among velar stops, such a worker will probably answer both questions without a $\lambda$. Now suppose the third worker is a native speaker of Hawaiian; since this language does not use voicing, such a worker will not be able to differentiate between $\{${\fransdvng k}$\}$ and $\{${\fransdvng g}$\}$, or between  $\{${\fransdvng K}$\}$ and $\{${\fransdvng G}$\}$.  Hence, it would be useful if this worker answered $\lambda$ for the question of differentiating among these two subchoices. 

The present paper allows workers to not respond, i.e.\ allowing a \emph{reject option}, as in the example.
Research in psychology suggests a greater tendency to select the reject option when the choice set offers several attractive alternatives but none that can be easily justified as the best, resulting in less arbitrary decisions \cite{dhar1997consumer}.  The reject option has previously been considered in the machine learning and signal processing literatures \cite{jung2014predicting,1054406, Bartlett:2008:CRO:1390681.1442792, doi:10.1287/deca.1080.0119, 5967817}, but we specifically consider worker behavior and aggregation rules for crowdsourcing with a reject option. To characterize performance, we derive a closed-form expression for the probability of a microtask being correct, together with the asymptotic performance when the crowd size is large.

Several methods have been proposed to deal with noisy crowd work when crowd workers are required to provide a definitive yes/no response \cite{hirth2013analyzing,VempatyVV2014,yue2014weighted,KargerOS2011b,KargerOS2011a,6891807,QuinnB2011,zhang2012reputation}, rather than allowing a reject option. Without the reject option, noisy responses to tasks cannot be tagged before aggregation so appropriate weights cannot be assigned \cite{yue2014weighted}. For instance, the popular majority voting rule weights all answers equally \cite{ruta2005classifier}, though new weighted aggregation rules have also been developed \cite{yue2014weighted,6891807}. We employed error-control codes and decoding algorithms to design reliable crowdsourcing systems with unreliable workers \cite{VempatyVV2014}.  A group control mechanism where worker reputation is used to partition the crowd into groups is presented in \cite{QuinnB2011, zhang2012reputation}; comparing group control to majority voting indicates majority voting is more cost-effective on less complex tasks \cite{hirth2013analyzing}.

Note that under typical crowdsourcing incentive schemes based on work volume, workers may respond to tasks for which they are not sufficiently skilled, even when a reject option is available.
We therefore also consider the case where a greedy fraction of the anonymous crowd workers complete all microtasks with random guesses to maximize their rewards.

The main contributions of this paper are threefold.
\begin{enumerate}
\item We study a new crowdsourcing paradigm where similarly-difficult microtasks are performed by workers having a reject option so they can skip microtasks if they believe they cannot respond reliably. We assign a weight for every worker's response based on the number of skipped microtasks and this gives rise to significant performance improvement when compared to when all workers must respond. We also provide asymptotic performance analysis as crowd size increases. Asymptotic behavior is insightful since crowdsourcing worker pools are typically large. 

\item In our weight assignment, a crowd worker's response is given more weight during aggregation if he/she responds to more microtasks. We also consider the setting where a fraction of greedy crowd workers respond to all tasks, first continuing to employ the oblivious strategy proposed above and investigating the error performance as a function of the fraction of greedy workers in the crowd. We then study another strategy, an expurgation strategy, where responses of workers that respond to all microtasks are discarded and the remaining workers' responses are assigned optimized weights.

\item An adaptive algorithm to switch between oblivious and expurgation strategies is derived, based on estimating several crowd parameters such as fraction of greedy workers in the crowd.
\end{enumerate}

Although the contributions listed above are stated for the crowdsourcing paradigm, our results hold for other signal classification tasks when decisions are made using signals that are quite uncertain. This is known as classification with a reject option \cite{HerbeiW2006} and has been focus of several recent studies in signal processing research including pattern recognition, image and speech classification \cite{1054406,Bartlett:2008:CRO:1390681.1442792,5967817,CondessaBCOK2013,PillaiFR2011}. More specifically, for a classification problem consisting of multiple (potentially unreliable) classifiers with a reject option, one can use the weighted aggregation rule where each classifier's weight is a function of its reject behavior and its reliability. Our analysis presented in this paper characterizes the asymptotic behavior of such a system. 

\section{Classification Task for Crowds}
Consider the situation where $W$ workers take part in an $M$-ary classification task. Each worker is asked $N$ simple binary questions, termed microtasks, which eventually lead to a decision among the $M$ classes. We assume it is possible to construct $N = \left\lceil {{{\log }_2}M} \right\rceil$ independent microtasks of equal difficulty. The workers submit results that are combined to give the final decision. A worker's answer to a single microtask is represented by either ``1'' (Yes) or ``0'' (No) \cite{VempatyVV2014,rocker2007paper} and so the $w$th worker's ordered answers ${\bf a}_w(i)$, $i\in \{ 1,2,\dots,N\}$ to all microtasks form an $N$-bit word, denoted ${\bf a}_w$.  

Each worker has the reject option of skipping microtasks; we denote a skipped answer as $\lambda$, whereas ``1/0'' (Yes/No) responses are termed \emph{definitive answers}. Due to variability in worker expertise, the probability of submitting definitive answers is different for different workers. Let $p_{w,i}$ represent the probability the $w$th worker submits $\lambda$ for the $i$th microtask. Similarly, let $\rho_{w,i}$ be the probability ${\bf a}_w(i)$, the $i$th answer of the $w$th worker, is correct given a definitive answer has been submitted. Due to worker anonymity, we study performance when $p_{w,i}$ and $\rho_{w,i}$ are realizations of certain probability distributions, denoted by $F_P(p)$ and $F_{\rho}(\rho)$, respectively. The corresponding means are denoted $m$ and $\mu$. Let $H_0$ and $H_1$ denote hypotheses corresponding to bits ``0'' and ``1'' for a single microtask, respectively. For simplicity of performance analysis, ``0'' and ``1'' are assumed equiprobable for every microtask. The crowdsourcing platform or fusion center (FC) collects the $N$-bit words from $W$ workers and aggregates results, as discussed next. 

%Let $\rho^{s,t}_{w,i}, s,t\in \{0,1\}$ denote the probability that ${\bf a}_w(i)=t$ under hypothesis $H_s$ given that $w$th worker has a definitive answer for the corresponding task.  and we assume $\rho^{s,t}_{w,i}=\rho^{1-s,t}_{w,i}$. Therefore, the correct probability $\rho_{w,i} = {\rho ^{1,1}_{w,i}} = {\rho ^{0,0}_{w,i}}$.

\subsection{Weighted Majority Voting}
We first investigate weighted majority voting as the fusion rule for classification. Let us consider all object classes as elements in the set $D=\{e_j, j=1,2,\dots,M\}$, where $e_j$ represents the $j$th class. As indicated earlier, a worker's definitive responses to the microtasks determine a subset in the original set $D$, consisting of multiple elements/classes. If all the responses from the crowd are definitive, the final subsets are singletons and a single class is identified as the object class. Since in our framework here, some microtasks may be answered with a response $\lambda$, the resulting subsets in this case will not be singletons and each element of the same corresponding subsets will have equal probability to be selected as the classification decision. Let us denote the subset determined by the definitive answers of the $w$th worker as $D_w\in D$. The task manager assigns the same weight to all elements in $D_w$ based on the $w$th worker's answer. With the submitted answers from $W$ workers, we determine the overall weight assigned to any $j$th element of $D$ as 
\begin{align}
\mathbb{W}\left( e_j \right) = \sum\limits_{w= 1}^W {{W_w}I_{{D_w}}\langle e_j \rangle } , j=1,2,\dots ,M, \ \ D_w\in D,
\end{align}
where $W_w$ is the weight\footnote{The assignment of these weights will be discussed later in the paper.} assigned to $D_w$, and $I_{D_w}\langle e_j \rangle$ is an indicator function which equals 1 if $e_j \in D_w$ and 0 otherwise.
Then the element $e_D$ with the highest weight is selected as the final class, and the classification rule is stated as
\begin{align}\label{3}
e_D= \arg \mathop {\max }\limits_{{{e_j} \in {D}} }\left\{ {\mathbb{W}\left( e_j \right)} \right\}.
\end{align}
Notice that conventional majority voting requires full completion of all microtasks without the reject option and has identical $W_w$ for each worker's decision.

Next, we show how the problem formulated in \eqref{3} can be further decomposed.
\begin{proposition}
The classification rule in \eqref{3} is equivalent to a bit-by-bit decision rule as the $i$th bit, $i=1,\ldots, N$, is decided by
\begin{align}\label{4}
{{\sum\limits_{w=1} ^{W}{W_w} {I_1}\left\langle {i,w} \right\rangle }}\overset{{{H}}_1}{\underset{{{H}}_0}{\gtrless}}{{\sum\limits_{w=1}^{W} {W_w} I_0\left\langle {i,w} \right\rangle }},
\end{align}
where ${I_s}\left\langle {i,w} \right\rangle $, $s\in\{0,1\}$, is the indicator function which is 1 if the $w$th worker's answer to the $i$th bit is ``$s$'', otherwise ${I_s}\left\langle {i,w} \right\rangle =0$. For tie-breaking, randomly choose 0 or 1.
\end{proposition}
\begin{IEEEproof}
The class $e_D$ corresponds to a unique $N$-bit word. Thus, if the $i$th bit of the $N$-bit word corresponding to the class $e_D$ is equal to $s$, $s$ has the same weight as assigned to $e_D$, which is greater than or equal to the symbol $1-s$. This gives the decision rule stated in \eqref{4}.
\end{IEEEproof}

\subsection{Optimal Bit-by-bit Bayesian Aggregation}
Let ${\mathcal A(i)} = [{{\bf a}_1}(i),{{\bf a}_2}(i), \ldots ,{{\bf a}_W}(i)]$ denote all the answers to $i$th microtask collected from the crowd. For the binary hypothesis testing problem corresponding to the $i$th bit of the $N$-bit word, the log-likelihood ratio test is
\begin{align}\label{OBAC}
\log \frac{{P\left( {{H_1}|{\mathcal A(i)}} \right)}}{{P\left( {{H_0}|{\mathcal A(i)}} \right)}}\overset{{{H}}_1}{\underset{{{H}}_0}{\gtrless}}0.
\end{align}
We can express the likelihood ratio as
\begin{small}
\begin{align}
\frac{{P\left( {{H_1}|{\mathcal A(i)}} \right)}}{{P\left( {{H_0}|{\mathcal A(i)}} \right)}} &= \frac{{\prod\limits_{w = 1}^W {P\left( {{\bf a}_w(i)|{H_1}} \right)} }}{{\prod\limits_{w = 1}^W {P\left( {{\bf a}_w(i)|{H_0}} \right)} }} \nonumber\\
&= \frac{{\prod\limits_{{S_1}} {\left( {1 - {p_{w,i}}} \right){\rho _{w,i}}} \prod\limits_{{S_0}} {\left( {1 - {p_{w,i}}} \right)\left( {1 - {\rho _{w,i}}} \right)\prod\limits_{{S_\lambda }} {{p_{w,i}}} } }}{{\prod\limits_{{S_1}} {\left( {1 - {p_{w,i}}} \right)\left( {1 - {\rho _{w,i}}} \right)} \prod\limits_{{S_0}} {\left( {1 - {p_{w,i}}} \right){\rho _{w,i}}\prod\limits_{{S_\lambda }} {{p_{w,i}}} } }},
\end{align}
\end{small}
where $S_1$ is the set of $w$ such that ${\bf a}_w(i)=1$, $S_0$ is the set of $w$ such that ${\bf a}_w(i)=0$ and $S_\lambda$ is the set of $w$ such that ${\bf a}_w(i)=\lambda$, respectively.
Then, it is straightforward to show that the test for the decision on the $i$th bit is
\begin{align}
\sum\limits_{{w\in S_1}} {\log \frac{{{\rho _{w,i}}}}{{1 - {\rho _{w,i}}}}} \overset{{{H}}_1}{\underset{{{H}}_0}{\gtrless}}\sum\limits_{{w\in S_0}} {\log \frac{{{\rho _{w,i}}}}{{1 - {\rho _{w,i}}}}}.
\end{align}
Note that the optimal Bayesian criterion can also be viewed as the general weighted majority voting rule in \eqref{4} with weight $W_w={\log \frac{{{\rho _{w,i}}}}{{1 - {\rho _{w,i}}}}}$, which is also called the Chair-Varshney rule \cite{ChairV1986}. Note that \eqref{4} represents majority voting when $W_w=1$.

However, this optimal Bayesian criterion can only be used if $\rho_{w,i}$ for every worker is known \textit{a priori}, which is usually not available since the crowd involved is anonymous and thus it is not possible to extract prior information such as $\rho_{w,i}$ from the answers they submit. The difficulty in obtaining prior information makes the simple majority voting scheme very effective and therefore widely used \cite{ruta2005classifier}. We will show later in the paper that our proposed method can be employed in practical situations and outperforms conventional majority voting. Estimation of $\rho_{w,i}$ is not needed while the mean of $\rho_{w,i}$, $\mu$, is estimated in a practical way.

\subsection{Class-based Aggregation Rule}
For the general weighted majority voting scheme where $e_C$ denotes the correct class, the probability of misclassification is
\begin{align}
{P_m} &= \Pr \left( {{e_D} \ne {e_C}} \right) \nonumber\\
&= \Pr \left( {\arg \mathop {\max }\limits_{{e_j} \in D} \left\{ {\sum\limits_{w = 1}^W {{W_w}I_{D_w}\left\langle {{e_j}} \right\rangle } } \right\} \ne {e_C}} \right)\nonumber\\
& = 1 - \Pr \left( {\arg \mathop {\max }\limits_{{e_j} \in D} \left\{ {\sum\limits_{w = 1}^W {{W_w}I_{D_w}\left\langle {{e_j}} \right\rangle } } \right\} = {e_C}} \right).
\end{align}

A closed-form expression for the error probability $P_m$ cannot be derived without an explicit expression for $W_w$; hence it is difficult to determine the optimal weights to minimize $P_m$.

Consequently, we consider an optimization problem based on a different objective function and propose a novel weighted majority voting method that outperforms simple majority voting. Note that $e_D$ is chosen as the decision for classification such that $e_D$ has the maximum overall weight collected from all the workers. Thus, we maximize the average overall weight assigned to the correct class while the overall weight collected by all the elements remains the same as the other existing methods such as majority voting. We state the optimization problem over the weights as
\begin{equation}\label{max}
\begin{array}{l}
\text{maximize}\ \ {E_C}\left[ {{\mathbb{W}}} \right]\\
\text{subject to}\ \ {E_O}\left[ {{\mathbb{W}}} \right] = {K}
\end{array}
\end{equation}
where ${E_C}\left[ {{\mathbb{W}}} \right]$ denotes the crowd's average weight contribution to the correct class and ${E_O}\left[ {{\mathbb{W}}} \right]$ denotes the average weight contribution to all the possible classes. $ {K}$ is set to a constant so that we are looking for a maximized portion of weight contribution to the correct class while the weight contribution to all the classes remains fixed. This procedure ensures that one can not obtain greater $ {E_C}\left[ {{\mathbb{W}}} \right]$ by simply increasing the weight for each worker's answer, while $K$ results in a normalized weight assignment scheme. If two weight assignment schemes share the same value of ${E_O}\left[ {{\mathbb{W}}} \right]$, one can expect better performance by the scheme with higher $ {E_C}\left[ {{\mathbb{W}}} \right]$. Thus, $K$ facilitates a relatively easier performance comparison of different weight assignment schemes.

\section{Crowdsourcing System With Honest Crowd Workers}
We first consider the case where the crowd is entirely composed of honest workers, which means that the workers are not greedy and honestly observe, think, and answer the questions corresponding to microtasks, and skip a question that they are not confident about. The $w$th worker responds with a $\lambda$ to the $i$th microtask with probability $p_{w,i}$. Next, we derive the optimal weight $W_w$ for the $w$th worker in this case.
\begin{proposition}
To maximize the normalized average weight assigned to the correct classification element, the weight for $w$th worker's answer is given by
\begin{align}
W_w={\mu}^{-n},
\end{align}
where $\mu=E[\rho_{w,i}]$ and $n$ is the number of definitive answers that the $w$th worker submits.
\end{proposition}

\begin{IEEEproof}
See Appendix A.
\end{IEEEproof}

Note that in previous works on crowdsourcing, workers were forced to make a hard decision for every single bit, which is the case when the weight derived above is set to an identical value. Here the weight depends on the number of questions answered by a worker. In fact, if more questions are answered, the weight assigned to the corresponding worker's answer is larger. In our crowdsourcing framework, if the worker believes a definitive answer can be generated, we assume that the corresponding correct probability is greater than one half. Then, a larger number of definitive answers results in a greater chance that the quality of the worker is higher than others. Increased weight can put more emphasis on the contribution of high-quality workers in that sense and improve overall classification performance.
\subsection{Estimation of $\mu$}
Before the proposed aggregation rule can be used, note that $\mu$ has to be estimated to assign the weight for every worker's answers. Here, we give two approaches to estimate $\mu$. 
\subsubsection{Training-based} In addition to the $N$ microtasks, the task manager inserts additional questions to estimate the value of $\mu$ of the crowd. The answers to such ``gold standard'' questions are of course known to the manager \cite{difallah2012mechanical,jung2015modeling}. By checking the crowd worker's answers, $\mu$ can be estimated. Suppose that the first $T$ questions are training questions, followed by $N$ questions for classification. Let $\bar {\bf B}$ be the $T$-bit correct answers to the training questions that the manager has. First, we calculate the ratio $r(w)$ as
\begin{align}
r(w)=\sum\limits_{i=1}^{T}\frac{{I_{{\bar {\bf B}}(i)}\left\langle {{\bf a}_w(i)} \right\rangle }}{I(w)},
\end{align}
where $I_x\langle  y \rangle$ is the indicator function which equal 1 if $x=y$ and 0 otherwise, and $I( w ) = \sum_{i = 1}^T {\left( {{I_1}\langle i,w \rangle  + {I_0}\langle {i,w} \rangle } \right)}$. In order to avoid the cases where some workers submit $\lambda$ for all the training questions, we estimate $\mu$ as follows
\begin{align}
\label{eq:mu_est}
\hat \mu=\frac{1}{W-\epsilon}\sum\limits_{w=1}^{W}{r(w)},
\end{align}
where $\epsilon$ is the number of workers that submit all $\lambda$ for the training questions and the corresponding $r(w)$ is set to $0$.

\subsubsection{Majority-voting based} We use majority voting to obtain the initial aggregation result and set it as the benchmark to estimate $\mu$. First, all the answers ${\bf a}_w(i)$ are collected to obtain the benchmark ${\mathcal B}(i)$ by traditional majority voting, where $i=1,\ldots , N$. Note that ${\mathcal B}(i)$ may contain $\lambda$ since it is possible that all answers ${\bf a}_w(i)$ have $\lambda$ at the same position. Then, for the $w$th worker, we calculate the ratio $r(w)$ as
\begin{align}
r(w) = \sum\limits_{i = 1}^N {\frac{{I_{{\mathcal{B}}(i)}\left\langle {{{\bf{a}}_w}(i)} \right\rangle }}{{I\left( w \right)}}} ,
\end{align}
where we set $I_{ \lambda}\langle \lambda \rangle =0$, and $I(w) = \sum_{i = 1}^N {\left( {{I_1}\langle {i,w} \rangle  + {I_0}\langle {i,w} \rangle } \right)}$. As before, we estimate $\mu$ as 
in \eqref{eq:mu_est}, but where $\epsilon$ is the number of workers that submit $\lambda$ for all microtasks.

\subsection{Performance Analysis}
In this subsection, we characterize performance of the proposed classification framework in terms of probability of correct classification $P_c$. Note that we have overall correct classification only when all the bits are classified correctly.\footnote{When $N>\log_2M$, the $N$-bit answer after aggregation may correspond to a class that does not exist; this is also misclassification.}

First, we restate the bit decision criterion in \eqref{4} as
\begin{align}\label{test}
{{\sum\limits_{w=1} ^{W}{T_w}  }}\overset{{{H}}_1}{\underset{{{H}}_0}{\gtrless}}{{0 }}\end{align}
with $T_w={W_w}\left( {{I_1}\langle {i,w} \rangle  - {I_0}\langle {i,w} \rangle } \right)$,
where the resulting 
$T_w\in \{-\mu^{-N},-\mu^{-N+1},\ldots, -\mu^{-1},0,\mu^1,\ldots,\mu^{N-1},\mu^N\}$.
\begin{proposition}
For the $i$th bit, the probability mass function of $T_w$ under hypothesis $H_s$, $\Pr \left( {{T_w}|{H_s}} \right)$, for $s\in \{0,1\}$, is:
\begin{align}\label{tw}
&\Pr \left( {T_w}=I(-1)^{t+1}\mu^{-n}|{H_s} \right) \nonumber\\
&= \left\{ {\begin{array}{*{20}{c}}
{\rho _{w,i}^{1 - \left| {s - t} \right|}{\left( {1 - {\rho _{w,i}}} \right)^{\left| {s - t} \right|}}\varphi_n(w), I = 1}\\
{p_{w,i},\ \ \ \ \ \ \ \ \ \ \ \ \ \ \ \ \ \ \ \ \ \ \ \ \ \ \ \ \ \  I = 0}
\end{array}} \right.,\nonumber\\
&t\in \{0,1\}, n\in\{1,\ldots,N\},
\end{align}
where $I = {I_1}\langle {i,w} \rangle  + {I_0}\langle {i,w} \rangle $, $\varphi_n(w) =(1-p_{w,i})\sum_C {\prod_{j = 1\hfill\atop
j \ne i\hfill}^N {p_{w,j}^{{k_j}}{{\left( {1 - {p_{w,j}}} \right)}^{1 - {k_j}}}} }$ and $C$ is the set
\begin{align}
C = \left\{ {\left\{ {{k_1},{k_2}, \ldots ,{k_{i - 1}},{k_{i + 1}}, \ldots ,{k_N}} \right\}: \sum\limits_{j = 1\hfill\atop
j \ne i\hfill}^N {{k_j} = N - n} } \right\}
\end{align}
with ${k_j} \in \left\{ {0,1} \right\}$.
\end{proposition}
\begin{IEEEproof}
See Appendix B.
\end{IEEEproof}

Since hypotheses $H_0$ and $H_1$ are assumed equiprobable, the correct classification probability for the $i$th bit $P_{c,i}$ is
\begin{align}
{P_{c,i}} = \frac{{1 + {P_{d,i}} - {P_{f,i}}}}{2},
\end{align}
where $P_{d,i}$ is the probability of deciding the $i$th bit as ``1'' when the true bit is ``1'' and $P_{f,i}$ is the probability of deciding the $i$th bit as ``1'' when the true bit is ``0''. 

\begin{proposition}\label{pci}
The probability of correct classification for the $i$th bit $P_{c,i}$ is
\begin{align}
{P_{c,i}} = \frac{1}{2}&+ \frac{1}{2}\sum\limits_S {\binom{W}{\mathbb{Q}}} \left( {F_i\left( \mathbb{Q} \right) - F_i^{\prime}\left( \mathbb{Q} \right)} \right) \nonumber\\
&+ \frac{1}{4}\sum\limits_{S^\prime} {\binom{W}{\mathbb{Q}}} \left( {F_i\left( \mathbb{Q} \right) - F_i^{\prime} \left( \mathbb{Q} \right)} \right)
\end{align}
with
\begin{align}
{ F}_i({\mathbb Q}) = \prod\limits_{w \in {G_{{\lambda}}}} {{p_{w,i}}}  \prod\limits_{w \in {G_{{0}}}} {\left( {1 - {\rho _{w,i}}} \right)} {\varphi _n}\left( w \right)  \prod\limits_{w \in {G_{{1}}}} {{\rho _{w,i}}} {\varphi _n}\left( w \right)
\end{align}
and
\begin{align}
{ F_i}^\prime({\mathbb Q}) = \prod\limits_{w \in {G_{{\lambda}}}} {{p_{w,i}}}  \prod\limits_{w \in {G_{{1}}}} {\left( {1 - {\rho _{w,i}}} \right)} {\varphi _n}\left( w \right)\prod\limits_{w \in {G_{0}}} {{\rho _{w,i}}} {\varphi _n}\left( w \right),
\end{align}
where
\begin{align}
\mathbb{Q}=\left\{({{q_{ - {N}}},{q_{ - {{N +1}}}}, \ldots {q_{{N}}}}):  \sum\limits_{n = -N}^N{{q_{{n}}}  = W}\right\}
\end{align} with natural numbers $q_n$ and $q_0$, $G_0$ denotes the worker group that submits ``0'' for $i$th microtask, $G_1$ the group that submits ``1'' and $G_{\lambda}$ the group that submits $\lambda$, and
\begin{align}
{S} = \left\{ {\mathbb{Q}: } {\sum\limits_{n = 1}^N {{{\mu}^{-n}}\left( {{q_{{n}}} - {q_{ - {n}}}} \right)}  > 0}\right\},
\end{align} 
\begin{align}
S^\prime = \left\{ {\mathbb{Q}:\sum\limits_{n = 1}^N {{\mu ^{ - n}}\left( {{q_n} - {q_{ - n}}} \right)}  = 0} \right\},
\end{align}
 and $\binom{W}{\mathbb{Q}} = \frac{{W!}}{{\prod\limits_{n =  - N}^N {{q_n}!} }}$.
\end{proposition}
\begin{IEEEproof}
See Appendix C.
\end{IEEEproof}

\begin{proposition}
The probability of correct classification $P_c$ in the crowdsourcing system is
\begin{align}
P_c=\Big[\frac{1}{2}&+ \frac{1}{2}\sum\limits_S {\binom{W}{\mathbb{Q}}} \left( {F\left( \mathbb{Q} \right) - F^{\prime}\left( \mathbb{Q} \right)} \right) \nonumber\\
&+ \frac{1}{4}\sum\limits_{S^\prime} {\binom{W}{\mathbb{Q}}} \left( {F\left( \mathbb{Q} \right) - F^{\prime} \left( \mathbb{Q} \right)} \right)\Big]^N,
\end{align}
where
\begin{small}
\begin{align}\label{fq}
F({\mathbb{Q}}) = {m^{{q_0}}}\prod\limits_{n = 1}^N {{{\left( {1 - \mu } \right)}^{{q_{ - n}}}}{\mu ^{{q_n}}}{{\left( {C_{N - 1}^{n - 1}{{\left( {1 - m} \right)}^n}{m^{N - n}}} \right)}^{{q_{ - n}} + {q_n}}}} 
\end{align}
\end{small}
and
\begin{small}
\begin{align}\label{f'q}
F^{\prime}({\mathbb{Q}}) = {m^{{q_0}}}\prod\limits_{n = 1}^N {{{\left( {1 - \mu } \right)}^{{q_n}}}{\mu ^{{q_{ - n}}}}{{\left( {C_{N - 1}^{n - 1}{{\left( {1 - m} \right)}^n}{m^{N - n}}} \right)}^{{q_{ - n}} + {q_n}}}} ,
\end{align}
\end{small}
\end{proposition}
\begin{IEEEproof}
See Appendix D.
\end{IEEEproof}

In practice, the number of workers for the crowdsourcing task is large (in the hundreds). Thus it is of value to investigate the asymptotic system performance when $W$ increases without bound.
\begin{proposition}
As the number of workers $W$ approaches infinity, the probability of correct classification $P_c$ can be expressed as
\begin{align}\label{pc}
P_c=\left[Q\left( { - \frac{{{M}}}{{\sqrt {{V{{}}}} }}} \right)\right]^N,
\end{align}
where $Q(x) = \frac{1}{{\sqrt {2\pi } }}\int_x^{\infty}  {e^{\frac{{ - t^2 }}{2}} dt}$, and $M$ and $V$ are given as
\begin{align}
M={\frac{{W\left( {2\mu  - 1} \right)\left( {1 - m} \right)}}{\mu }} {\left( {\frac{1}{\mu } - \left( {\frac{1}{\mu } - 1} \right)m} \right)^{N - 1}},
\end{align}
and
\begin{align}
V=\frac{{W\left( {1 - m} \right)}}{{{\mu ^2}}}{\left( {\frac{1}{{{\mu ^2}}} - \left( {\frac{1}{{{\mu ^2}}} - 1} \right)m} \right)^{N - 1}} - \frac{{{M^2}}}{W}.
\end{align}
\end{proposition}
\begin{IEEEproof}
See Appendix E.
\end{IEEEproof}
For large but finite crowds, the asymptotic result \eqref{pc} is a good characterization of actual performance. Let us therefore consider \eqref{pc} in more detail.
First, we rewrite \eqref{pc} as
\begin{align}\label{qf}
P_c=\left[Q\left( { - \sqrt {\frac{W}{{\frac{1}{{f\left( {\mu ,m} \right)}} - 1}}} } \right)\right]^N,
\end{align}
where
\begin{align}\label{f}
f\left( {\mu ,m} \right) = \left( {1 - m} \right){\left( {2\mu  - 1} \right)^2}{\left( {g\left( {\mu ,m} \right)} \right)^{N - 1}},
\end{align}
and
\begin{align}\label{g}
g\left( {\mu ,m} \right) = \frac{{{{\left( {1 - \left( {1 - \mu } \right)m} \right)}^2}}}{{1 - \left( {1 - {\mu ^2}} \right)m}}.
\end{align}
\begin{theorem}
The probability of correct classification in the crowdsourcing system increases with increasing size of the crowd $W$.
\end{theorem}
\begin{IEEEproof}
Follows from \eqref{qf} as the probability of correct classification increases monotonically with respect to $W$. 
\end{IEEEproof}

\begin{theorem}
The probability of correct classification in the crowdsourcing system increases with increasing $\mu$.
\end{theorem}
\begin{IEEEproof}
We take the partial derivative of $g(\mu,m)$ with respect to $\mu$ and obtain
\begin{align}
\frac{{\partial g}}{{\partial \mu }} = \frac{{2m(1 - \mu )\left( {1 - m} \right){\mathbf A} }}{{{{\mathbf B} ^2}}},
\end{align}
where ${\mathbf A}=m\mu-m+1$, and
${\mathbf B}=m\mu^2-m+1$.

Clearly $\tfrac{{\partial g}}{{\partial \mu}} >0$. Recall \eqref{qf}, \eqref{f}, and \eqref{g}: a larger $P_c$ results as $\mu$ increases. Then, the  classification performance of the task in the crowdsourcing system also increases.
\end{IEEEproof}

To obtain the relation between crowd's performance in terms of $P_c$ and $m$,
we take the partial derivative of $f(\mu,m)$ with respect to $m$ and obtain
\begin{small}
\begin{align}
&\frac{1}{(2\mu-1)^2}\frac{{\partial f}}{{\partial m}}\nonumber\\
& =  \left( {N - 1} \right)\left( {1 - m} \right)\left( {\frac{{{{\mathbf A} ^2}{{\left( {\mu^2-1 } \right)}}}}{{{{\mathbf B} ^2}}} + \frac{{2{\mathbf A} \left( {1 - \mu } \right)}}{{\mathbf B} }} \right){\left( {\frac{{{{\mathbf A} ^2}}}{{\mathbf B} }} \right)^{N - 2}} \nonumber\\
&\ \ \ - {\left( {\frac{{{{\mathbf A} ^2}}}{{\mathbf B} }} \right)^{N - 1}}.
\end{align}
\end{small}

After some mathematical manipulations, we observe:
\begin{itemize}
\item
When $m>\frac{1}{1+\mu}$, we can guarantee that $\frac{{\partial f}}{{\partial m}}<0$, which means that the crowd performs better as $P_c$ increases with decreasing $m$.

\item
When $m<\frac{1}{1+\mu}$ and $N \ge \frac{{{{\left( {m\mu  - m + 1} \right)}^2}}}{{\left( {1 - m} \right){{\left( {1 - \mu } \right)}^2}\left( {m\mu  + m - 1} \right)}} + 1$, we can guarantee that $\frac{{\partial f}}{{\partial m}}>0$, which means that the crowd performs better as $P_c$ increases with increasing $m$.
\end{itemize}

These two observations indicate that a larger probability of the crowd responding to the $i$th microtask with $\lambda$ does not necessarily degrade crowd's performance in terms of the detection of the $i$th microtask. 

This counterintuitive result follows since even though the crowd skips more microtasks, the optimized weight takes advantage of the number of unanswered questions and extracts more information. For this to happen, the number of microtasks $N$ has to be greater than a lower limit. Since a larger $N$ induces more diversity in the number of unanswered questions, the existence of the lower limit means that this diversity can actually benefit the performance using the proposed scheme.

\subsection{Simulation Results}
In this subsection, we compare the performance of the proposed crowdsourcing system where crowd workers are allowed to skip microtasks with the conventional majority voting method in a hard-decision fashion, which means workers are forced to make a decision even if the workers believe that no definitive answers could be provided. The number of equiprobable classes is set as $M=8$.

\begin{figure}[hbp]
   \centering
   \includegraphics[width=3.4in]{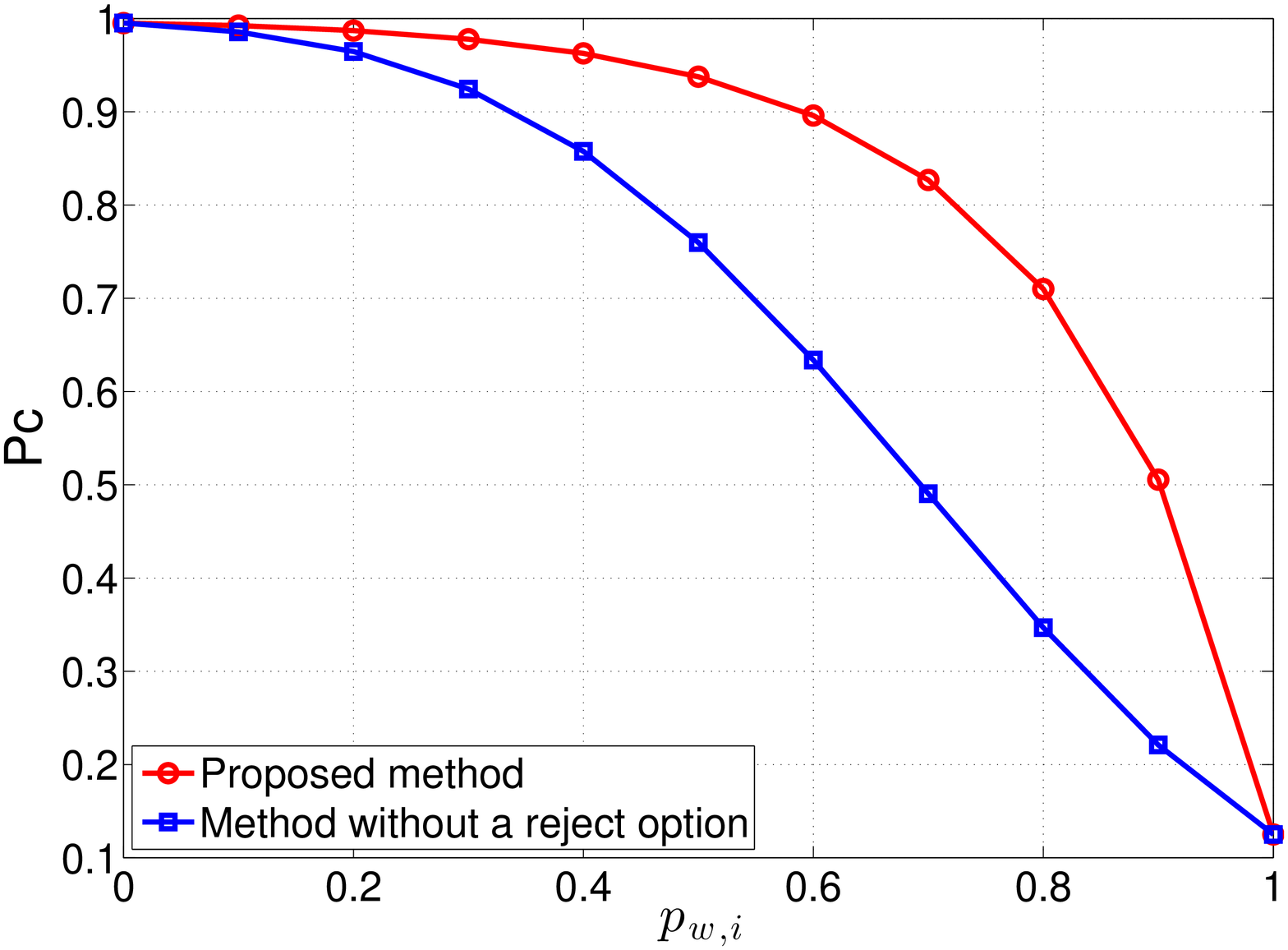} % requires the graphicx package
   \caption{Proposed approach compared to majority voting at $\rho_{w,i}=0.8$.}
   \label{figA}
\end{figure}
\begin{figure}[h]
   \centering
   \includegraphics[width=3.4in]{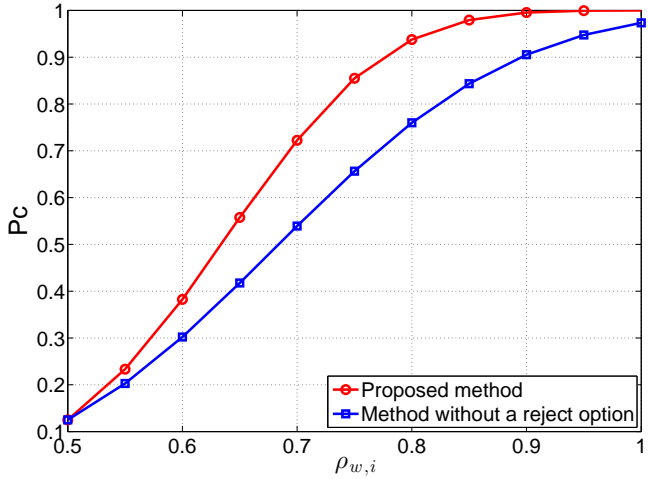} % requires the graphicx package
   \caption{Proposed approach compared to majority voting at $p_{w,i}=0.5$.}
   \label{figB}
\end{figure}

Fig.~\ref{figA} compares performance when $W=20$ workers take part in the task. We consider here that workers have a fixed $p_{w,i}$ for each microtask and $\rho_{w,i}=0.8$. We observe that performance degrades as $p_{w,i}$ gets larger, which means that the workers have a higher probability of not submitting an answer to the microtask. A remarkable performance improvement associated with our proposed approach is observed. The two curves converge at $p_{w,i}=1$ with $P_c$ being equal to 0.125. At this point, with the majority-based approach, each worker gives random answers for each microtask whereas the workers using the proposed scheme skip all the questions and the tie-breaking criterion is used to pick a random bit for every microtask. In Fig.~\ref{figB}, we fix $p_{w,i}=0.5$ and vary $\rho_{w,i}$ to compare the resulting $P_c$. Notable performance improvement is also seen. The point at $\rho_{w,i}=0.5$ indicates that the worker is making a random guess even if he/she believes that he/she can complete the corresponding microtask correctly. The performance improves as $\rho_{w,i}$ gets larger, which means that the crowd is able to give higher-quality definitive answers.

\begin{figure}[h]
   \centering
   \includegraphics[width=3.4in]{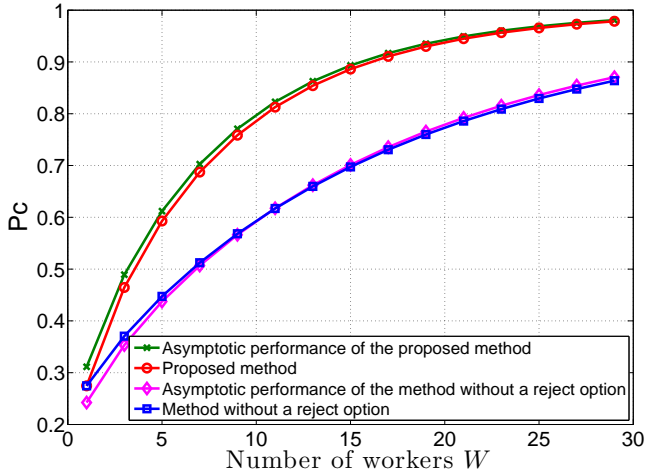} % requires the graphicx package
   \caption{Proposed approach compared to majority voting with various worker sizes at $p_{w,i} \sim U(0,1)$ and $\rho_{w,i} \sim U(0.6,1)$.}
   \label{figC}
\end{figure}

\begin{figure}[h]
   \centering
   \subfigure[Performance comparison.]{
   \includegraphics[width=3.3in]{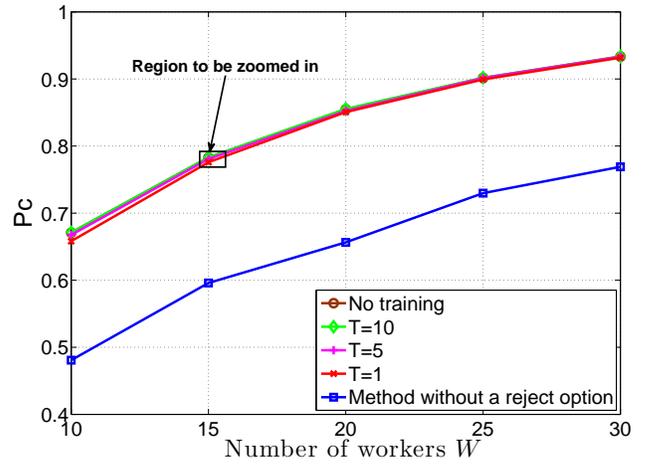}
      \label{figDa}
   } \\% requires the graphicx package\\
   \subfigure[Zoomed-in version]{
   \includegraphics[width=3.3in]{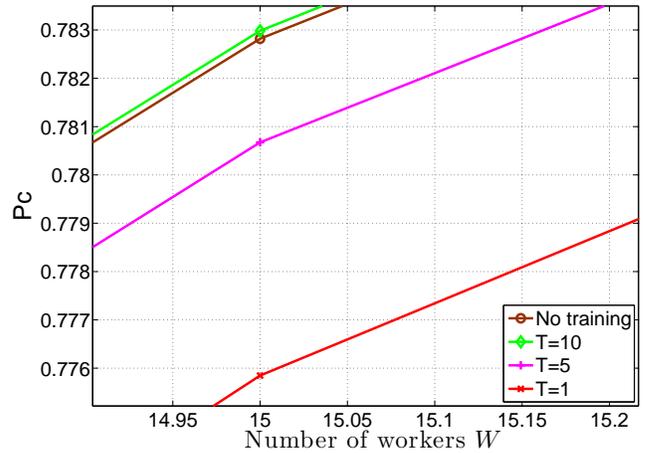}
     \label{figDb}
   }
   \caption{Proposed approach compared to majority voting with various worker sizes at $p_{w,i} \sim U(0,1)$ and $\rho \sim U(0.5,1)$. Two methods are used to estimate $\mu$ for weight assignment. One uses training to insert $T$ additional microtasks for estimation, whereas the other one uses the decision results of majority voting as a benchmark to estimate $\mu$. (a) provides the performance comparison while (b) is a zoomed-in region which is indicated in the box in (a).}
   \label{figD}
\end{figure}

In Fig.~\ref{figC}, we compare performance with different number of workers, also showing the asymptotic performance characterization. Here, we consider different qualities of the individuals in the crowd which is represented by variable $p_{w,i}$ with uniform distribution $U(0,1)$ and $\rho_{w,i}$ with $U(0.6,1)$. First, it is observed that a larger crowd completes the classification task with higher quality. The asymptotic curves are derived under the assumption of a very large crowd, which are the bounds on the performance of the systems. It is not a very difficult task to derive that the asymptotic performance for conventional majority voting is 
\begin{align}
P_c=\left[Q\left( -\sqrt{\frac{W^2(2l-1)}{4l-4l^2}} \right)\right]^N\nonumber,
\end{align}
where $l=\mu+m(0.5-\mu)$. Therefore, the asymptotic gap in the performance between conventional majority voting and the proposed method can also be obtained.
We find that the asymptotic curves are quite a tight match to the actual performance. Again, we can see a significant improvement in $P_c$ brought by the proposed approach.

We now include the estimation of $\mu$ in Fig.~\ref{figD} for weight assignment. Observe in Fig.~\ref{figDa} that the performance improves as the number of workers increases and also observe that the proposed approach is significantly better than majority voting. Second, the performance of the proposed approach is significantly better than that of traditional majority voting, and it changes for different estimation settings. As is expected, a larger number of training questions result in higher performance of the system as observed in Fig. \ref{figDb}. Another interesting finding is that the performance with training based estimation can exceed that with majority voting as a benchmark only when a relatively large number of training questions are used. We can see from the figure that the ``Training'' method with $T=10$ slightly outperforms the method without training and based on ``Majority-voting''. However, the number of microtasks $N$ is only 3, which is much smaller than the training size. Quite a bit of extra overhead besides the classification task will need to be added if the training method is adopted. Hence, it is reasonable to employ ``Majority-voting'' method together with the proposed approach for the classification task with crowdsourcing.

\begin{figure}[h]
   \centering
   \includegraphics[width=3.4in]{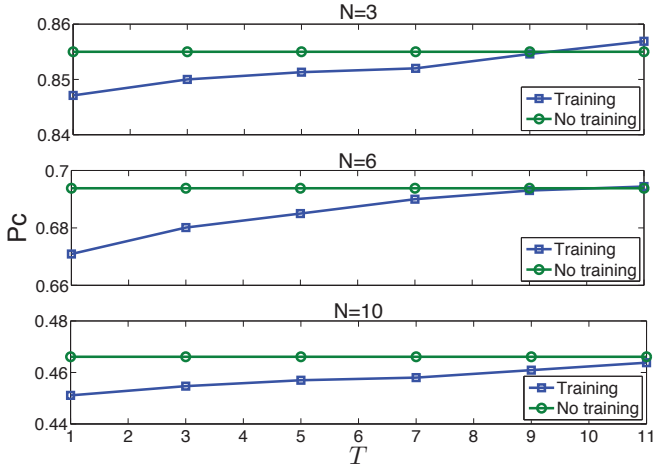} % requires the graphicx package
   \caption{Performance vs overhead tradeoff. The crowd size is set as $W=20$ and $N=3$, 6, and 10, from top to bottom, respectively. }
   \label{overhead}
\end{figure}

Fig.~\ref{overhead} shows the performance as a function of overhead with different numbers of microtasks, to illustrate the performance gap between the two methods. Observe that the method without training exhibits remarkable advantage in probability of correct classification $P_c$ when a reasonable number of additional microtasks are inserted. Improved performance of the training-based method is shown when $T$ gets larger as this results in more accurate estimation. To have comparable performance with the method without training, the training-based method requires even more additional microtasks when the original number of microtasks $N$ increases. With enough microtasks inserted, the training method can outperform the one without. Again, this result encourages employing the method without training in practice.

\section{Crowdsourcing System With Greedy Crowd Workers}
In the previous section, we considered conscientious crowd workers who respond only if they have confidence in their ability to respond. In our formulation, the weight assigned to a worker's response increases with the number of microtasks that the worker responds to and this contributes to the selection of the correct class thereby enhancing classification performance. In a reward-based system, such honest workers should be compensated and actually rewarded for completing as many tasks as possible. However, if there are workers in the crowd who are greedy and their goal is to obtain as much compensation without regard to the system goal of enhanced classification performance, such a reward mechanism can encourage greedy workers to randomly complete all microtasks to get a higher reward. This would result in a degradation in the classification performance of the system. Indeed, Mason and Watts observed that increasing financial incentives increases the quantity of work performed, but not necessarily the quality \cite{MasonW2009}. In this section, we study system performance when a part of the crowd completes all microtasks with random guesses. In other words, these ``greedy'' workers submit $N$-bit codewords, which are termed as full-length answers in the sequel. We investigate the crowdsourcing system with such workers by assuming the correct probability for each microtask is $1/2$ to represent the situation where the worker responds randomly to each microtask. 

Insertion of a gold standard question set is the most widely used method to address the issue of greedy workers. This, however, comes at the cost of a large question set to avoid workers spotting recurrent questions \cite{difallah2012mechanical}. Besides, this is not effective in practice since one of the fundamental reasons for crowdsourcing is to collect classified labels that we do not have \cite{jung2015modeling}. In this paper, we do not insert a gold standard question set and instead study two different strategies for this problem. The \emph{Oblivious Strategy} continues to use the same scheme we derived in the previous section ignoring the existence of greedy workers. In the \emph{Expurgation Strategy}, we discard the answers of workers who only give full-length answers, to reduce the impact of greedy workers on the overall system performance. Let $\alpha$ denote the fraction of greedy workers in the crowd. Note that in this strategy, we will also discard the responses of those honest workers that provided definitive answers to all microtasks. Note the greedy workers are not being punished in any way here; only in the aggregation strategy their responses are being ignored.

\subsection{Oblivious Strategy}
In this strategy, we continue to use the same weight allocation scheme as for honest workers, which can be expressed as
\begin{align}
{W_w} = {\alpha _1}{\mu ^{ - n}},
\end{align}
where the factor $\alpha_1$ is introduced to satisfy the constraint ${E_O}\left[ {{\mathbb{W}}} \right] = {K}$.

The average contribution from the crowd to the correct class and all the classes can be given respectively as
\begin{flalign}
{E_C}\left[ {{\mathbb{W}}} \right] = &\sum\limits_{w=1}^{W\alpha} {\alpha _1}{\mu ^{ - N}}\frac{1}{{{2^N}}} \nonumber\\
&+ \sum\limits_{w = W\alpha  + 1}^W {\sum\limits_{n = 0}^{N - 1} {{\alpha _1}{\mu ^{ - n}}C_N^n{{\left[ {\left( {1 - m} \right)\mu } \right]}^n}{m^{N - n}}} }\nonumber \\
 = &\frac{{W\alpha {\alpha _1}}}{{{{\left( {2\mu } \right)}^N}}} + {\alpha _1}W\left( {1 - \alpha } \right),
\end{flalign}
and
\begin{align}
{E_O} \left[ {{\mathbb{W}}} \right] =& {\sum\limits_{w = 1}^{W\alpha } {{\alpha _1}\left( {\frac{1}{\mu }} \right)} ^N} \nonumber\\&+ \sum\limits_{w = W\alpha  + 1}^W {\sum\limits_{n = 0}^{N - 1} {{\alpha _1}{\mu ^{ - n}}{2^{N - n}}C_N^n{{\left( {1 - m} \right)}^n}{m^{N - n}}} } \nonumber\\
 =&W\alpha {\alpha _1}{\left( {\frac{1}{\mu }} \right)^N} + \left( {W - W\alpha } \right){\alpha _1}{\left( {\frac{{1 - m}}{\mu } + 2m} \right)^N} \nonumber\\
 =&K.
 \end{align}
 
 Therefore, we can calculate $\alpha_1$ and obtain $E_C[\mathbb{W}]$ as:
\begin{align}
{\alpha _1} = \frac{K}{{W\alpha {{\left( {\frac{1}{\mu }} \right)}^N} + \left( {W - W\alpha } \right){{\left( {\frac{{1 - m}}{\mu } + 2m} \right)}^N}}},
\end{align}
\begin{align}
{E_C\left[\mathbb{W}\right]} = \frac{{{K}\alpha {{\left( {\frac{1}{{2\mu }}} \right)}^N} + {K}\left( {1 - \alpha } \right)}}{{\alpha {{\left( {\frac{1}{\mu }} \right)}^N} + \left( {1 - \alpha } \right){{\left( {\frac{{1 - m}}{\mu } + 2m} \right)}^N}}}.
\label{ec1}
\end{align}

\begin{proposition}\label{pf2}
The probability of correct classification $P_c$ when the Oblivious Strategy is used is 
\begin{align}
{P_c} =\Big[ \frac{1}{2} &+ \frac{1}{2}\sum\limits_S {\binom{W}{\mathbb{Q}_1}} \left( {F(\mathbb{Q}_1) - F^{\prime}(\mathbb{Q}_1)} \right) \nonumber\\&+\frac{1}{4}\sum\limits_{S^\prime} {\binom{W}{\mathbb{Q}_1}} \left( {F(\mathbb{Q}_1) - F^{\prime}(\mathbb{Q}_1)} \right)\Big] ^N
\end{align}
with
\begin{align}
&F({{\mathbb Q}_1}) \nonumber\\&= { {\frac{{m^{{q_0}}}}{2^{W\alpha }}}}\prod\limits_{n = 1}^N {{{\left( {1 - \mu } \right)}^{{q_{ - n}}}}{\mu ^{{q_n}}}{{\left( {C_{N - 1}^{n - 1}{{\left( {1 - m} \right)}^n}{m^{N - n}}} \right)}^{{q_{ - n}} + {q_n}}}} 
\end{align}
and
\begin{align}
&F^{\prime}({\mathbb Q}_1)\nonumber\\& = { {\frac{{m^{{q_0}}}}{2^{W\alpha }}}}\prod\limits_{n = 1}^N {{{\left( {1 - \mu } \right)}^{{q_n}}}{\mu ^{{q_{ - n}}}}{{\left( {C_{N - 1}^{n - 1}{{\left( {1 - m} \right)}^n}{m^{N - n}}} \right)}^{{q_{ - n}} + {q_n}}}} 
\end{align}
where
\begin{align}
{{\mathbb Q}_1} = \left\{ {({q_{ - N}},{q_{ - N + 1}}, \ldots {q_N}):\sum\limits_{n =  - N}^N {{q_n} = W - W\alpha } } \right\},
\end{align} with natural numbers $q_n$, 
\begin{align}
{S_1} = \left\{ {{{\mathbb Q}_1}:\sum\limits_{n = 1}^N {{\mu ^{ - n}}\left( {{q_n} - {q_{ - n}}} \right)}  + {\mu ^{ - N}}W\alpha  > 0} \right\},
\end{align}
\begin{align}
{S_1}^\prime  = \left\{ {{{\mathbb Q}_1}:\sum\limits_{n = 1}^N {{\mu ^{ - n}}\left( {{q_n} - {q_{ - n}}} \right)}  + {\mu ^{ - N}}W\alpha  = 0} \right\},
\end{align} and $\binom{W}{\mathbb{Q}_1} = \frac{{W!}}{{\prod_{n =  - N}^N {{q_n}!} }}$.
\end{proposition}
\begin{IEEEproof}
See Appendix F.
\end{IEEEproof}

\subsection{Expurgation Strategy}
In this strategy, all definitive answers of length of $N$ bits are discarded to avoid answers from greedy workers. The classification decision is made based on the answers with maximum number of bits equal to $N-1$. To proceed, we need the weight for every worker's answer in this case. We begin by restating the optimization problem
 \begin{equation}\label{max2}
\begin{array}{l}
\text{maximize}\ \ {E_C}\left[ {{\mathbb{W}}} \right]\\
\text{subject to}\ \ {E_O}\left[ {{\mathbb{W}}} \right] = {K}
\end{array}
\end{equation}
and we have
\begin{align}
{E_C}\left[ {{\mathbb{W}}} \right] &= \sum\limits_{w = 1}^{W - W\alpha } {\sum\limits_{n = 0}^{N - 1} {{W_w}\binom{N}{n}{{\left[ {\left( {1 - m} \right)\mu } \right]}^n}{m^{N - n}}} } \nonumber\\
 &= \sum\limits_{w = 1}^{W - W\alpha } {\sum\limits_{n = 0}^{N - 1} {{W_w}{\mu ^n}{x^{n - N}}{P_x}\left( n \right)} },
 \end{align}
 where 
 \begin{align}
{P_x}\left( n \right) = \binom{N}{n}{\left( {1 - m} \right)^n}{\left( {mx} \right)^{N - n}},
\end{align}
and $x$ is such that
\begin{align}\label{px}
\sum\limits_{n = 0}^{N - 1} {{P_x}\left( n \right)}  = 1.
\end{align}
Then, we can write
 \begin{align}
 {E_C}\left[ {{\mathbb{W}}} \right]
 \le \sum\limits_{w = 1}^{W - W\alpha } {\sqrt {\sum\limits_{n = 0}^{N - 1} {{{\left( {{W_w}{\mu ^n}{x^{n - N}}} \right)}^2}{P_x}\left( n \right)} } \sqrt {\sum\limits_{n = 0}^{N - 1} {{P_x}\left( n \right)} } },
 \end{align}
 and the equality holds only if 
 \begin{align}
{W_w}{\mu ^n}{x^{n - N}}\sqrt {{P_x}\left( n \right)}  = {\alpha _2}\sqrt {{P_x}\left( n \right)} ,
\end{align}
where the factor $\alpha_2$ is introduced to satisfy the constraint $E_O[\mathbb{W}] = K$.
Hence, we have the maximum of $E_C[W_w]$ as
 \begin{align}\label{ec2}
  {E_C}\left[ {{\mathbb{W}}} \right]= W\left( {1 - \alpha } \right){\alpha _2},
\end{align}
when
\begin{align}
{W_w}= {\alpha _2}{\mu ^{-n}}{x^{N-n}}.
\end{align}

To obtain the value of $x$, we rewrite \eqref{px} as:
\begin{align}
{\left( {1 - m + mx} \right)^N} - {\left( {1 - m} \right)^N} = 1,
\end{align}
and $x$ is given as
\begin{align}\label{xx}
x = \frac{{{{\left( {1 + {{\left( {1 - m} \right)}^N}} \right)}^{\frac{1}{N}}} + m - 1}}{m}.
\end{align}

For this strategy, the overall weight constraint is given as
\begin{small}
\begin{align}
{E_O\left[\mathbb{W}\right]} &= \sum\limits_{w = 1}^{W - W\alpha } {\sum\limits_{n = 0}^{N - 1} {{\alpha _2}{\mu ^{ - n}}{x^{N - n}}{2^{N - n}}\binom{N}{n}{{\left( {1 - m} \right)}^n}{m^{N - n}}} } \nonumber\\
 &= W\left( {1 - \alpha } \right){\alpha _2}\left[ {{{\left( {\frac{{1 - m}}{\mu } + 2mx} \right)}^N} - {{\left( {\frac{{1 - m}}{\mu }} \right)}^N}} \right] \nonumber\\
 &= \ K.\end{align}
 \end{small}
By substituting this result back into \eqref{ec2}, the maximum value of $E_C[\mathbb{W}]$ can be written as
\begin{align}
{E_C\left[\mathbb{W}\right]} = \frac{K}{{{{\left( {\frac{{1 - m}}{\mu } + 2mx} \right)}^N} - {{\left( {\frac{{1 - m}}{\mu }} \right)}^N}}}.
\label{ec22}
\end{align}

Note that the weight could be $W_w=\mu^{-n}x^{-n}$ when the Expurgation Strategy is employed in practice, where $x$ is given by \eqref{xx}.

\begin{proposition}\label{pf3}
The probability of correct classification $P_c$ when the Expurgation Strategy is used is
\begin{align}
{P_c} = \Big[\frac{1}{2} &+ \frac{1}{2}\sum\limits_S {\binom{W}{\mathbb{Q}_2}} \left( {F(\mathbb{Q}_2) - F^{\prime}(\mathbb{Q}_2)} \right) \nonumber\\&+\frac{1}{4}\sum\limits_{S^\prime} {\binom{W}{\mathbb{Q}_2}} \left( {F(\mathbb{Q}_2) - F^{\prime}(\mathbb{Q}_2)} \right)\Big]^N
\end{align}
with
\begin{small}
\begin{align}
F({{\mathbb Q}_2}) =
 {m^{{q_0}}}&\prod\limits_{n = 1}^{N - 1} {{{\left( {1 - \mu } \right)}^{{q_{ - n}}}}{\mu ^{{q_n}}}{{\left( {C_{N - 1}^{n - 1}{{\left( {1 - m} \right)}^n}{m^{N - n}}} \right)}^{{q_{ - n}} + {q_n}}}} 
\end{align}
\end{small}
and
\begin{small}
\begin{align}
F^{\prime}({{\mathbb Q}_{2}}) = 
{m^{{q_0}}}&\prod\limits_{n = 1}^{N - 1} {{{\left( {1 - \mu } \right)}^{{q_n}}}{\mu ^{{q_{ - n}}}}{{\left( {C_{N - 1}^{n - 1}{{\left( {1 - m} \right)}^n}{m^{N - n}}} \right)}^{{q_{ - n}} + {q_n}}}} ,
\end{align}
\end{small}
where
\begin{small}
\begin{align}
{{\mathbb Q}_2} = \left\{ {({q_{ - N+1}},{q_{ - N + 2}}, \ldots {q_{N-1}}):\sum\limits_{n =  - N + 1}^{N - 1} {{q_n} \le W - W\alpha } } \right\}
\end{align}
\end{small}
with natural numbers $q_n$, 
and
\begin{align}
{S_2} = \left\{ {{{\mathbb Q}_2}:\sum\limits_{n = 1}^{N - 1} {{\mu ^{ - n}}{x^{ - n}}\left( {{q_n} - {q_{ - n}}} \right)}  > 0} \right\},
\end{align} 
\begin{align}
{S_2}^\prime  = \left\{ {{{\mathbb Q}_2}:\sum\limits_{n = 1}^N {{\mu ^{ - n}}{x^{ - n}}\left( {{q_n} - {q_{ - n}}} \right)}  = 0} \right\} 
\end{align} and $\binom{W}{\mathbb{Q}_2} = \frac{{W!}}{{\prod_{n =  - N+1}^{N-1} {{q_n}!} }}$.

\end{proposition}
\begin{IEEEproof}
See Appendix G.
\end{IEEEproof}

\subsection{Adaptive Algorithm}
We now investigate the adaptive use of our two strategies to improve system performance. The goal is to find a threshold to distinguish when one strategy will outperform the other, so as to allow switching.

Note that the two strategies described in the previous subsections are associated with the same overall weight for all classes. Thus, we compare the crowd's total contribution to the correct class under this condition and derive the corresponding switching scheme. From \eqref{ec1} and \eqref{ec2}, this can be expressed in \eqref{switch},
\begin{figure*}[ht]
\normalsize
\begin{align}\label{switch}
\frac{{\alpha {K}{{\left( {\frac{1}{{2\mu }}} \right)}^N} + {K}(1 - \alpha) }}{{\alpha {{\left( {\frac{1}{\mu }} \right)}^N} + \left( {1 - \alpha } \right){{\left( {\frac{{1 - m}}{\mu } + 2m} \right)}^N}}}  
\overset{{\text {Oblivious Strategy}}}{\underset{{\text {Expurgation Strategy}}}{\gtrless}}
\frac{{K}}{{{{\left( {\frac{{1 - m}}{\mu } + 2mx} \right)}^N} - {{\left( {\frac{{1 - m}}{\mu }} \right)}^N}}},
\end{align}
\hrulefill \vspace*{4pt}
\end{figure*}
which can be simplified to have the switching threshold of $\alpha$ as
\begin{align}\label{sc}
\alpha \left( {{{\left( {\frac{1}{\mu }} \right)}^N} - {\gamma _1}{{\left( {\frac{1}{{2\mu }}} \right)}^N} - {\gamma _2} + {\gamma _1}} \right)
\overset{{\text {Expurgation Strategy}}}{\underset{{\text {Oblivious Strategy}}}{\gtrless}}
 {\gamma _1} - {\gamma _2},
\end{align}
where
\begin{align}
\gamma_1={{{\left( {\frac{{1 - m}}{\mu } + 2mx} \right)}^N} - {{\left( {\frac{{1 - m}}{\mu }} \right)}^N}},
\end{align}
and
\begin{align}
\gamma_2={\left( {\frac{{1 - m}}{\mu } + 2m} \right)^N}.
\end{align}

To obtain the threshold associated with $\alpha$ in the switching criterion \eqref{sc}, $\mu$ and $m$ should be estimated first. The previous section established a simple and effective method to estimate $\mu$ based on majority voting. Therefore, we again use majority voting to get initial detection results, which are then set as the benchmark to estimate $\mu$. Note that estimation of $\mu$ is based on the answers without the full-length ones to avoid degradation resulting from the greedy workers.

The performance of this integrated scheme can be derived using Propositions \ref{pf2}, \ref{pf3} and the switching criterion \eqref{sc}.

\subsection{Joint Estimation of $m$ and $\alpha$}
The threshold on $\alpha$ is specified based on the estimated values of $m$ and $\mu$. Then, we estimate $\alpha$ and compare it with the corresponding threshold to switch the strategies adaptively. Even though we discard the full-length answers and take advantage of the rest and estimate $m$, it is an inaccurate estimate because the discarded answers also contain those from the honest workers. 

Several works have studied the estimation of $\alpha$ in crowdsourcing systems \cite{allahbakhsh2013quality,hirth2013analyzing,zhang2012reputation}, which can be divided into two categories: one studies the behavior of the workers in comparison to the honest control group \cite{hirth2013analyzing}; the other one learns worker's reputation profile \cite{allahbakhsh2013quality,zhang2012reputation}, which is stored and updated over time to identify the greedy ones from the crowd. However, both categories of estimation methods are not suitable here due to the anonymous nature of crowd workers. The first category suffers from the difficulty in extracting the honest group from the anonymous crowd while the second requires identification of every worker.

Since the worker's quality is assumed to be i.i.d., we give a joint parametric estimation method of both $m$ and $\alpha$ based on maximum likelihood estimation (MLE).

As defined earlier, out of $W$ workers, $q_n+q_{-n}$ workers submit answers of $n$ bits, $0 \le n \le N$. Thus, the probability mass function of the number of submitted answers given $m$ and $\alpha$ is obtained in \eqref{mle}, 
\begin{figure*}[!ht]
\normalsize
\begin{align}\label{mle}
f\left( {{q_n+q_{-n}}|m,\alpha } \right) 
=
 \left\{ {\begin{array}{*{20}{c}}
{\binom{W - W\alpha }{{q_n+q_{-n}}}{A_{N,n,m}}^{{q_n+q_{-n}}}{{\left( {1 - {A_{N,n,m}}} \right)}^{W - W\alpha  - {q_n-q_{-n}}}},0 \le n < N}\\
{\binom{W - W\alpha }{{q_N+q_{-N}} - W\alpha }{{\left( {1 - m} \right)}^{N\left( {{q_N+q_{-N}} - W\alpha } \right)}}\left( {1 - {{\left( {1 - m} \right)}^N}} \right),n = N}
\end{array}} \right.
\end{align}
\hrulefill \vspace*{4pt}
\end{figure*}
where ${A_{N,n,m}} = \binom{N}{n}{\left( {1 - m} \right)^n}{m^{N - n}}$, is defined as the expectation of the probability of a single worker submitting $n$ definitive answers.

Because of the independence of workers, we can form the likelihood statistic as
\begin{align}
L\left( {m,\alpha } \right) = \sum\limits_{n = 0}^N {\log } f\left( {{q_n+q_{-n}}|m,\alpha } \right).
\end{align}
Therefore, the ML estimates of $m$ and $\alpha$, which are denoted by $\hat m$ and $\hat \alpha$, can be obtained as
\begin{align}
\left\{ {\hat m,\hat \alpha } \right\}=\arg \mathop {\max }\limits_{\left\{ {m,\alpha } \right\} \in \left[ {0,1} \right]} L\left( {m,\alpha } \right) .
\end{align}

Once we have $\hat \mu$, $\hat m$ and $\hat \alpha$, we can adaptively switch to the suitable strategy using \eqref{sc}.

\subsection{Simulation Results}
\begin{figure}[h] %  figure placement: here, top, bottom, or page
   \centering
   \includegraphics[width=3.4in]{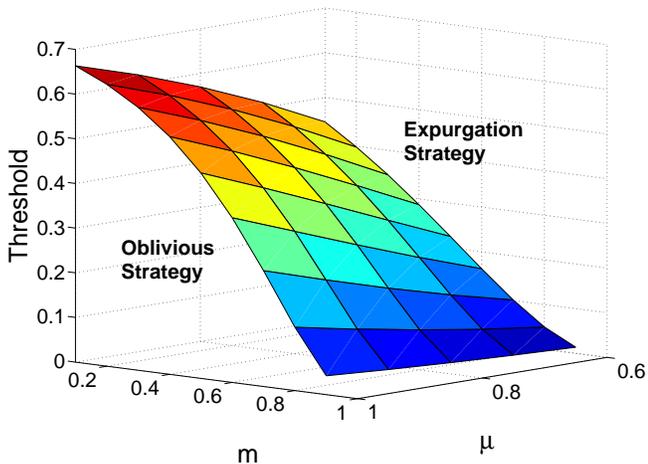} 
   \caption{Threshold to switch between strategies.}
   \label{fig:threshold}
\end{figure}

Now, we present some simulation results to illustrate the performance of our proposed algorithm. First, the theoretical value of the threshold for adaptive switching between the strategies is obtained for different values of $m$ and $\mu$ based on \eqref{sc}. We switch to the Expurgation Strategy if the fraction of greedy workers $\alpha$ is greater than the threshold. Otherwise we stick to the Oblivious Strategy. As we can observe from Figure \ref{fig:threshold}, when $m$ decreases and $\mu$ increases, which means the quality of the crowd improves, the threshold increases. This implies that the crowdsourcing system employing the Oblivious Strategy can tolerate a higher fraction of greedy workers in the crowd and, therefore, instead of discarding all the answers from the greedy workers and a part from the honest workers who submit full-length answers, it is better to keep them as long as the honest ones can perform well. The effect of greedy workers' answers can be compensated by the high-quality answers of the honest workers in the crowd.

Next, we give the estimation results for $\hat \alpha$ in Table \ref{my-label} using the proposed MLE method. The crowd quality parameters $p_{w,i}$ and $\rho_{w,i}$ are drawn from distributions $U(0,1)$ and $U(0.5,1)$ respectively. The number of microtasks $N$ and the number of workers $W$ are set to 3 and 20, respectively.

\begin{table}[h]
\centering
\caption{Estimation of $\alpha$}
\label{my-label}
\begin{tabular}{|l|l|l|l|l|l|l|l|l|l|}
\hline
$\alpha$      & 0.1  & 0.2   & 0.3  & 0.4   & 0.5  & 0.6   & 0.7  & 0.8   & 0.9  \\ \hline
$\hat \alpha$ & 0.11 & 0.26 & 0.36 & 0.48 & 0.58 & 0.67 & 0.79 & 0.87 & 0.96 \\ \hline
\end{tabular}
\end{table}
\begin{figure}[h] %  figure placement: here, top, bottom, or page
   \centering
   \includegraphics[width=3.4in]{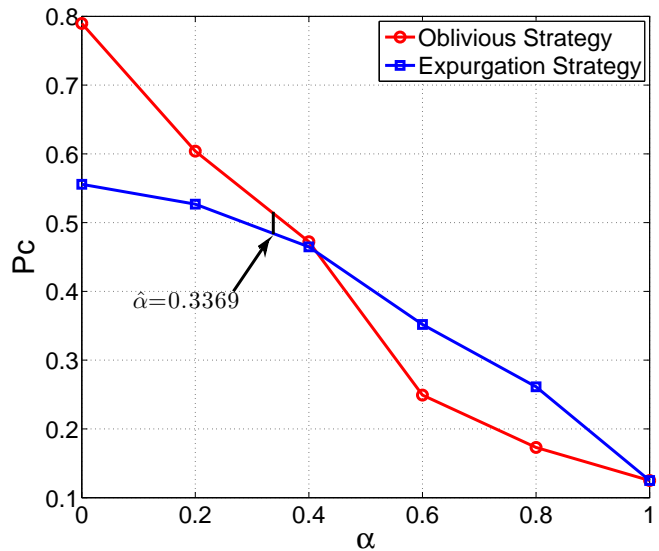} 
   \caption{Performance of both the strategies with greedy workers}
   \label{fig:strategy}
\end{figure}
In Fig. \ref{fig:strategy}, the performance of the proposed adaptive scheme is shown. The system parameters are the same as in previous simulations except that the crowd size $W$ is set equal to 15. The crowdsourcing system starts with the estimation of the parameters $\mu$, $m$, and $\alpha$. Once it has obtained $\hat \mu$ and $\hat m$, the system calculates the threshold value and compares it with $\hat \alpha$, and then decides the strategy to be used. Next, the system allocates weights to the answers for aggregation based on the strategy selected and makes the final classification decision. In Fig. \ref{fig:strategy}, we present the performance of both the strategies and the estimated threshold for switching is also presented. The performance deterioration caused by greedy workers is quite obvious as the probability of correct classification $P_c$ decreases for both strategies with increasing $\alpha$. The intersection of curves illustrates the need for strategy switching. The estimated Pareto frontier $\hat \alpha$ for switching is 0.3369 in this system setting, which is indicated in the figure by a line segment and is very close to the intersection of the two curves. Therefore, the actual performance curve of the proposed algorithm consists of the curve with squares when $\alpha<\hat \alpha$ and curve with circles when $\alpha>\hat \alpha$.

\section{Conclusion and Discussion}
We have studied a novel framework for crowdsourcing of classification tasks that arise in human-based signal processing, where an individual worker has the reject option and can skip a microtask if he/she has no definitive answer. We presented an aggregation approach using a weighted majority voting rule, where each worker's response is assigned an optimized weight to maximize the crowd's classification performance. We have shown our proposed approach significantly outperforms traditional majority voting and provided asymptotic performance characterization as an upper bound on performance. Further, we considered greedy workers in the crowd. An oblivious and an expurgation strategy were studied to deal with greed, and an algorithm to adaptively switch between the two strategies, based on the estimated fraction of greedy workers in the anonymous crowd, was developed to combat performance degradation.

We assumed in this paper that the workers' qualities are identically distributed. In some cases, it is possible that the workers' qualities are not identically distributed, which makes estimating the greedy fraction $\alpha$ difficult. The difficulty of microtasks might not be equal, which makes the microtask design quite challenging. Therefore, further research directions include the development of a general model to characterize the crowd quality, design of a robust method to estimate the fraction of greedy workers in the crowd, and binary microtask design with a reject option.

\appendices
\section{}\label{appB}
To solve problem \eqref{max}, we need $E_C[\mathbb{W}]$ and $E_O[\mathbb{W}]$. First, the $w$th worker can have weight contribution to $E_C[\mathbb{W}]$ only if all his/her definitive answers are correct. Thus, we have the average weight assigned to the correct element as
\begin{align}
E_C[\mathbb{W}] &= E_{p,\rho}\left[ {\sum\limits_{w = 1}^W {\sum\limits_{n = 0}^N {{W_w}P\left( {n,N - n} \right)} } } \right] \nonumber\\
&= \sum\limits_{w = 1}^W {E_{p,\rho}\left[ {\sum\limits_{n = 0}^N {{W_w}P\left( {n,N - n} \right)} } \right]} ,
\end{align}
where ${P\left( {n,N - n} \right)}$ represents the probability of $N-n$ bits equal to $\lambda$ and the rest of the $n$ definitive answers in  the $N$-bit word are correct.

Then, given a known $w$th worker, i.e., $p_{w,i}$ is known, we write
\begin{align}\label{5}
A_w(p_{w,i})=E_{\rho}\left[ {\sum\limits_{n = 0}^N {{W_w}P\left( {n,N - n} \right)|p_{w,i} } } \right].
\end{align}

Let $P_{\lambda}(n)$ denote the probability of the $w$th worker submitting $n$ definitive answers out of $N$ microtasks which only depends on $p_{w,i}$. Note that ${\sum_{n = 0}^N {P_{\lambda}(n)}  }=1$, and then \eqref{5} is upper-bounded using Cauchy-Schwarz inequality as follows:
\begin{align}
{A_w}(p_{w,i})& = \sum\limits_{n = 0}^N {E_\rho \left[ {{W_w}\rho \left( n \right)} \right]\sqrt {{P_\lambda }(n)} } \sqrt {{P_\lambda }(n)}  \nonumber \\ \label{8}& \le \sqrt {\sum\limits_{n = 0}^N {{{{E_\rho^2}\left[ {{W_w}\rho \left( n \right)} \right]}}{P_{\lambda}(n)} } } \sqrt {\sum\limits_{n = 0}^N {P_{\lambda}(n)}  } \\
& \triangleq \alpha_w(p_{w,i}),
\label{9}\end{align}
where $\rho(n)$ is the product of any $n$ out of $N$ variables $\rho_{w,i}$ as $i=1,2, \dots ,N$, and $\alpha_w$ is a positive quantity independent of $n$, which might be a function of $p_{w,i}$.
Note that equality holds in \eqref{8} only if 
\begin{align}
{{{E_{\rho}\left[ {{W_w}\rho \left( n \right)} \right]}\sqrt {P_{\lambda}(n)}  }}{{{}}} = \alpha_w(p_{w,i}) \sqrt {P_{\lambda}(n)} , 
\end{align}
which results in \eqref{9} and
\begin{align}\label{11}
{E_\rho\left[ {{W_w}\rho \left( n \right)} \right]}=\alpha_w(p_{w,i}).
\end{align}

Then we maximize the crowd's average weight corresponding to the correct class under the constraint $\int_{p_{w,i}} \Pr (p_{w,i} = x )dx=1$, and the maximization problem is written as
\begin{align}
 A &= E_p[A_w(p_{w,i})]=\int\limits_{p_{w,i}} {{\alpha _w(p_{w,i})}\Pr \left( {p_{w,i}= x} \right)dx}\nonumber\\
& \label{13} \le \sqrt {\int\limits_{p_{w,i}} {\alpha _w^2(p_{w,i})\Pr \left( {p_{w,i} = x} \right)dx} } \sqrt {\int\limits_{p_{w,i}} {\Pr \left( {p_{w,i} = x} \right)dx} } \\
& \triangleq \beta.
\end{align}
The equality \eqref{13} holds only if 
\begin{align}
\alpha_w(p_{w,i}) \sqrt {\Pr \left( {p_{w,i} = x}\right)}=\beta \sqrt {\Pr \left( {p_{w,i} = x}\right)},
\end{align}
with $\beta$ is a positive constant independent of $p_{w,i}$, and we conclude that $\alpha_w$ is also a positive quantity independent of $p_{w,i}$.
Then from \eqref{11}, we have
\begin{align}
{E_\rho\left[ {{W_w}\rho \left( n \right)} \right]}=\beta.
\end{align}
Since $\rho(n)$ is the product of $n$ variables, its distribution is not known \textit{a priori}. A possible solution to weight assignment is a deterministic value given by $W_wE[\rho(n)]=\beta$ and, therefore, we can write the weight as 
\begin{align}
W_w=\frac{\beta}{\mu^{n}}.
\end{align}

Then, we can express the crowd's average weight contribution to all the classes defined in \eqref{max} as
\begin{align}
{E_O}\left[ {{\mathbb{W}}} \right] &= \sum\limits_{w=1}^{W}E_{p,\rho}\left[ {\sum\limits_{n = 0}^N {\beta {\mu ^{ - n}}{2^{N - n}}{P_\lambda }\left( n \right)} } \right] \nonumber\\
&= \sum\limits_{w=1}^{W}\sum\limits_{n = 0}^N {\beta {\mu ^{ - n}}{2^{N - n}}\binom{N}{n}{{\left( {1 - m} \right)}^n}{m^{N - n}}}  \nonumber\\
&=W \beta {\left( {\frac{{1 - m}}{\mu } + 2m} \right)^N}=K.
\label{beta}
\end{align}
Thus, $\beta$ can be obtained from \eqref{beta} and we can obtain the weight by solving optimization problem \eqref{max} to get:
\begin{align}
W_w=\frac{{K}}{W{{\mu ^n}{{\left( {\frac{{1 - m}}{\mu } + 2m} \right)}^N}}}.
\end{align}
Note that the weight derived above has a term that is common for every worker. Since the voting scheme is based on comparison, we can ignore this factor and have the normalized weight as
\begin{align}
W_w={\mu}^{-n}.
\end{align}

\section{}\label{appC}
Note that 
\begin{align}{{{T}}_w} \in \{ { - {\mu ^{ - N}}, - {\mu ^{ - N + 1}}, \ldots , - {\mu ^{ - 1}},0,{\mu ^{ - 1}}, \ldots ,{\mu ^{ - N + 1}},{\mu ^{ - N}}} \},
\end{align} 
which can be written as
\begin{align}
{{T_w} = I{{\left( { - 1} \right)}^{t + 1}}{{\mu }^{ - n}}}, t\in \{0,1\}, I\in\{0,1\}, n\in\{1,\ldots,N\},
\end{align}
and leads to
\begin{align}
 &\Pr \left( {T_w} = I{{\left( { - 1} \right)}^{t + 1}}{{\mu }^{ - n}}|{H_s}\right)\nonumber\\
 &= \left\{ {\begin{array}{*{20}{c}}
{\Pr \left(T_w=\frac{ {{{\left( { - 1} \right)}^{t + 1}} } }{{{\mu }^{ n}}}|{H_s}\right), I = 1}\\
{\Pr \left( {{T_w} = 0|{H_s}} \right),  \ \ \ \ \ \ \ \ I = 0}
\end{array}} \right..
\end{align}
These two terms can be expressed as
\begin{align}
&\Pr \left( {{T_w} = \frac{{{{\left( { - 1} \right)}^{t + 1}}}}{{{{\mu }^n}}}|{H_s}} \right) \nonumber\\
&= \Pr \left( {{{\bf a}_w({i})} = t|{H_s}} \right)
\cdot  {{P}_\lambda}\left( { n|{{\bf a}_w({i})} = t,{H_s}} \right)\nonumber\\
 &= \rho _{w,i}^{1 - \left| {s - t} \right|}{\left( {1 - {\rho _{w,i}}} \right)^{\left| {s - t} \right|}}\sum\limits_C {\prod\limits_{j = 1\hfill\atop
j \ne i\hfill}^N {p_{w,j}^{{k_j}}{{\left( {1 - {p_{w,j}}} \right)}^{1 - {k_j}}}} } ,
\end{align}
and
\begin{align}
\Pr \left( {{T_w} = 0|{H_s}} \right)=p_{w,i}.
\end{align}

\section{}\label{appD}
Let $q_n, -N\le n\le N$, represent the number of workers that submit $|{n}|$ total definitive answers to all the microtasks. Specifically,  $n<0$ indicates the group of workers that submit ``0'' for the $i$th bit while $n>0$ indicates ``1''. For $n=0$, $q_0$ represents the number of workers that submit $\lambda$ for the $i$th bit. Since the workers independently complete the microtasks,  recalling the results in \eqref{tw}, the probabilities of the crowd's answer profile for the $i$th bit $\{G_0, G_1,G_{\lambda}\}$ can be obtained under $H_1$ and $H_0$ given $p_{w,i}$ and $\rho_{w,i}$ are expressed by ${ F}_i({\mathbb Q}) $ and ${ F}_i^\prime({\mathbb Q}) $, respectively.
Thus, $P_{d,i}$ given $p_{w,i}$ and $\rho_{w,i}$ can be expressed as
\begin{align}
{P_{d,i}} = \sum\limits_S \binom{W}{\mathbb{Q}} { F_i}(\mathbb{Q}) + \frac{1}{2}\sum\limits_{S^\prime} \binom{W}{\mathbb{Q}}{ F_i}(\mathbb{Q}) ,
\end{align}
where the first term on the right-hand side corresponds to the case where the aggregation rule gives a result of ``1'' and the second term indicates the case where ``1'' is given due to the tie-breaking of the aggregation rule.

Similarly, we can obtain $P_f$ given  $p_{w,i}$ and $\rho_{w,i}$ as
\begin{align}
{P_{f,i}} = \sum\limits_S \binom{W}{\mathbb{Q}} { F_i}^\prime({\mathbb Q}) + \frac{1}{2}\sum\limits_{S^\prime} \binom{W}{\mathbb{Q}}{ F_i}^\prime({\mathbb Q}).
\end{align}

Then, it is straightforward to obtain the desired result.

\section{}\label{appE}
We can have a correct classification if and only if all the bits are classified correctly. Thus, the expected probability of correct classification is given as
\begin{align}
P_c=E\left[ {\prod \limits _{i=1}^N P_{c,i}} \right],
\end{align}
which can be expressed, due to the independence of the microtasks, as
\begin{align}
P_c= \prod \limits _{i=1}^N E\left[ {P_{c,i}} \right].
\end{align}
Recall $P_{c,i}$ from Proposition \ref{pci}, and we can obtain:
\begin{align}
E\left[ {P_{c,i}} \right]=\frac{1}{2}&+ \frac{1}{2}\sum\limits_S {\binom{W}{\mathbb{Q}}} \left( {F\left( \mathbb{Q} \right) - F^{\prime}\left( \mathbb{Q} \right)} \right) \nonumber\\
&+ \frac{1}{4}\sum\limits_{S^\prime} {\binom{W}{\mathbb{Q}}} \left( {F\left( \mathbb{Q} \right) - F^{\prime} \left( \mathbb{Q} \right)} \right)
\end{align}
with $ F\left( \mathbb{Q} \right) $ and $F^{\prime} \left( \mathbb{Q} \right) $ defined in \eqref{fq} and \eqref{f'q}.
Thus we have the desired result.

\section{}\label{appF}
When $W$ goes to infinity, we show that $E\left[ {P_{c,i}} \right]=Q\left( { - \frac{{{M{{}}}}}{{\sqrt {{V{{}}}} }}} \right)$ and the desired result can be obtained. Based on the Central Limit Theorem \cite{deMoivre1756}, the test statistic in \eqref{test} is approximately Gaussian if $W\rightarrow \infty$ :
\begin{equation}
\sum\limits_{w = 1}^W {{T_w}} \sim \left\{ {\begin{array}{*{20}{c}}
{{\cal N}\left( {{M_1},{V_1}} \right),}&{{H_1}}\\
{{\cal N}\left( {{M_0},{V_0}} \right),}&{{H_0}}
\end{array}} \right.,
\end{equation}
where $M_s$ and $V_s$ are the means and variances of the statistic $\sum\limits_{w = 1}^W {{T_w}}$ under hypotheses $H_s$, respectively.

For the $w$th worker, we have the expectation of $T_w$ as
\begin{align}
{M_{{H_1}}} = \sum\limits_{t = 0}^1 {\sum\limits_{n = 1}^N {{{\left( { - 1} \right)}^{t + 1}}{\mu ^{ - n}}{\left( {{\rho _{w,i}}} \right)^{t }}{\left( {1 - {\rho _{w,i}}} \right)^{1-t}}{\varphi _n(w)}} } .
\end{align}
We define $M_1$ as
\begin{align}
M_1 &\triangleq WE\left[ {{M_{{H_1}}}} \right] \nonumber\\
&= W\sum\limits_{n = 1}^N {{\mu ^{ - n}}\left( {2\mu  - 1} \right)\binom{N-1}{n-1}{{\left( {1 - m} \right)}^n}{m^{N - n}} }\nonumber\\
&= {\frac{{W\left( {2\mu  - 1} \right)\left( {1 - m} \right)}}{\mu }} {\left( {\frac{1}{\mu } - \left( {\frac{1}{\mu } - 1} \right)m} \right)^{N - 1}}.
\end{align}
Likewise, we define $V_1$ as
\begin{align}
V_1 &\triangleq W\left( {E\left[ {T_w^2} \right] - {E^2}\left[ {{M_{{H_1}}}} \right]} \right) \nonumber\\
&= WE\left[\sum\limits_{t = 0}^1 {\sum\limits_{n = 1}^N {{\mu ^{ - 2n}}{\left( {{\rho _{w,i}}} \right)^{t}}{\left( {1 - {\rho _{w,i}}} \right)^{1-t}}{\varphi _n(w)}} } \right] - \frac{{{M_1^2}}}{W}\nonumber\\
&= \frac{{W\left( {1 - m} \right)}}{{{\mu ^2}}}{\left( {\frac{1}{{{\mu ^2}}} - \left( {\frac{1}{{{\mu ^2}}} - 1} \right)m} \right)^{N - 1}} - \frac{{{M_1^2}}}{W}.
\end{align}

Similarly, we can derive 
\begin{align}
M \triangleq {M_1} =  - {M_0},
\end{align}
and
\begin{align}
V \triangleq{V_1} = {V_0}.
\end{align}

By looking back at the decision criterion for the $i$th bit \eqref{test}, we can obtain the desired result.

\section{}\label{appG}
Since the workers complete the microtasks independently,  recall the results in \eqref{tw} and we have
\begin{align}
E\left[{P_{d,i}}\right] = \sum\limits_S \binom{W}{\mathbb{Q}_1} F(\mathbb{Q}_1) + \frac{1}{2}\sum\limits_{S^\prime} \binom{W}{\mathbb{Q}_1}F(\mathbb{Q}_1) ,
\end{align}
and
\begin{align}
E\left[{P_{f,i}}\right] = \sum\limits_S \binom{W}{\mathbb{Q}_1} F^{\prime}({\mathbb Q_1}) 
 + \frac{1}{2}\sum\limits_{S^\prime} \binom{W}{\mathbb{Q}_1}F^{\prime}({\mathbb Q}_1)  ,
\end{align}
with $F({\mathbb{Q}_1})$ and $F^{\prime}({\mathbb{Q}_1})$ as given above. Then, it is straightforward to obtain the desired result.

\section{}\label{appH}
Since the workers complete the microtasks independently,  recalling the results in \eqref{tw} we have
\begin{align}
E\left[{P_{d,i}}\right] = \sum\limits_S \binom{W}{\mathbb{Q}_2} F(\mathbb{Q}_2) + \frac{1}{2}\sum\limits_{S^\prime} \binom{W}{\mathbb{Q}_2}F(\mathbb{Q}_2) ,
\end{align}
and
\begin{align}
E\left[{P_{f,i}}\right] = \sum\limits_S \binom{W}{\mathbb{Q}_2} F^{\prime}({{\mathbb Q}_2}) 
 + \frac{1}{2}\sum\limits_{S^\prime} \binom{W}{\mathbb{Q}_2}F^{\prime}({\mathbb Q}_2)  
\end{align}
with $F({\mathbb{Q}_2})$ and $F^{\prime}({\mathbb{Q}_2})$ given above. Then, it is straightforward to obtain the desired result.

\bibliographystyle{IEEEtran}
\bibliography{abrv,conf_abrv,IEEEabrv,ref_Lqw}

\end{document}